\pdfoutput=1

\documentclass[11pt]{article}

\usepackage[]{ACL2023}

\usepackage{times}
\usepackage{latexsym}

\usepackage[T1]{fontenc}

\usepackage[utf8]{inputenc}

\usepackage{microtype}

\usepackage{inconsolata}

\usepackage{graphicx}
\usepackage{gb4e}\noautomath
\usepackage{enumitem}
\usepackage{amssymb}

\newcommand\blfootnote[1]{%
  \begingroup
  \renewcommand\thefootnote{}\footnote{#1}%
  \addtocounter{footnote}{-1}%
  \endgroup
}

\title{Morphological Inflection: A Reality Check}

\author{Jordan Kodner$^1$~~~Sarah Payne$^1$*~~~Salam Khalifa$^1$* \and Zoey Liu$^2$ \\
  $^1$Stony Brook University, 
  Dept. of Linguistics \& Institute for Advanced Computational Science\\
 $^2$University of Florida, 
  Dept. of Linguistics\\
\texttt{first.last@stonybrook.edu} \and \texttt{liu.ying@ufl.edu}}

\newcommand{\both}{\texttt{both}}
\newcommand{\feats}{\texttt{featsOnly}}
\newcommand{\lemma}{\texttt{lemmaOnly}}
\newcommand{\neither}{\texttt{neither}}
\newcommand{\featsAttested}{\texttt{featsAttested}}
\newcommand{\featsNovel}{\texttt{featsNovel}}

\newcommand{\uniform}{\textsc{Uniform}}
\newcommand{\weighted}{\textsc{Weighted}}
\newcommand{\aware}{\textsc{OverlapAware}}

\newcommand{\wuetal}{\textsc{chr-trm}}
\newcommand{\cluzhbfour}{\textsc{cluzh-b4}}
\newcommand{\cluzhgr}{\textsc{cluzh-gr}}
\newcommand{\nonneural}{\textsc{nonneur}}

\begin{document}
\maketitle
\begin{abstract}
Morphological inflection is a popular task in sub-word NLP with both practical and cognitive applications. For years now, state-of-the-art systems have reported high, but also highly variable, performance across data sets and languages. We investigate the causes of this high performance and high variability; we find several aspects of data set creation and evaluation which systematically inflate performance and  obfuscate differences between languages. To improve generalizability and reliability of results, we propose new data sampling and evaluation strategies that better reflect likely use-cases. Using these new strategies, we make new observations on the generalization abilities of current inflection systems.

\end{abstract}

\section{Introduction}

\blfootnote{*Denotes equal contribution}Morphological inflection is a task with wide-reaching applications in NLP, linguistics, and cognitive science. As the reverse of lemmatization, it is a critical part of natural language generation, particularly for languages with elaborate morphological systems \citep{bender-2009-linguistically,oflazer2018turkish}. Since morphological inflection is a particular type of well-defined regular string-to-string mapping problem \citep{roark2007computational,chandlee2017computational}, it is also useful for testing the properties of different neural network architectures.
Within cognitive science and linguistics, computational models of inflection have a long history in arbitrating between competing theories of morphological representation and acquisition \citep[surveyed in][]{pinker2002past,Seidenberg2014QuasiregularityAI}, and inflection is often a focus of computational typology \cite{bjerva-augenstein-2018-phonology,elsner2019modeling}.

However, despite the task's popularity, standard evaluation practices have significant weaknesses.
We discuss three aspects of these practices which hamper investigators' ability to derive informative conclusions. \textbf{(1)} Uniform sampling, which creates unnatural train-test splits, \textbf{(2)} Evaluation of single data splits, which yields unstable model rankings, 
and \textbf{(3)} uncontrolled overlaps between train and test data components, which obscure diagnostic information about systems' ability to perform morphological generalizations.


\begin{figure}[t]
    \centering
    \includegraphics[width=2.6in]{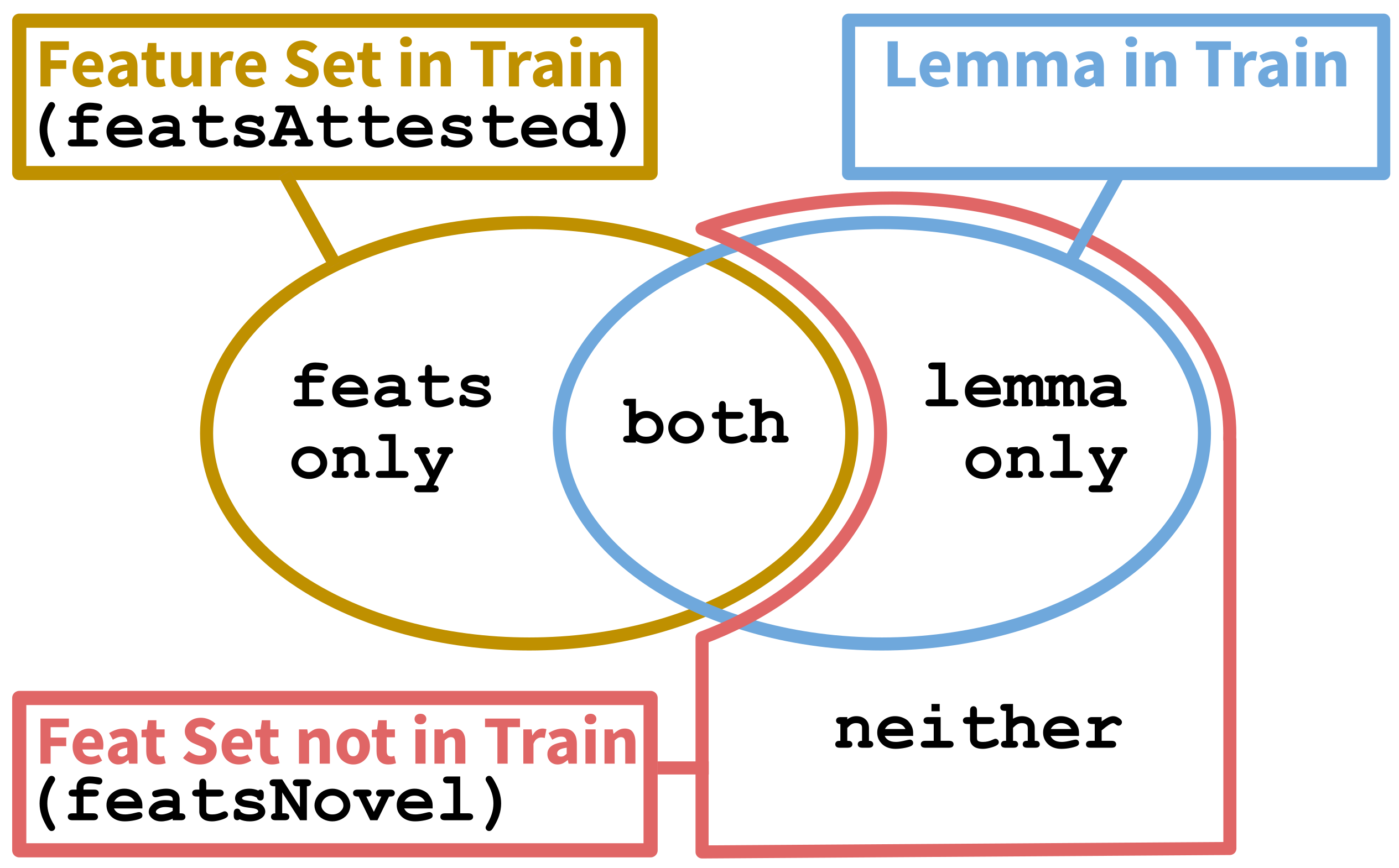}
    \caption{The four logically possible train-eval overlap types if evaluation data consists of (\texttt{lemma}, \texttt{feature set}) pairs: \both, \feats, \lemma, \neither, as well as \featsAttested = \both\ $\cup$ \feats\ and \featsNovel = \lemma\ $\cup$ \neither.\vspace{-0.5cm}}
    \label{fig:overlap_diagram}
\end{figure}

\subsection{Practice 1: Uniform Sampling}

Training and evaluation sets have been (with some exceptions) sampled uniformly by type from a corpus such as those available in the UniMorph Database \citep{kirov-etal-2018-unimorph,mccarthy-etal-2020-unimorph,batsuren-etal-2022-unimorph}. While practical to implement for corpora that lack frequency information, uniform sampling is also unrealistic because morphological forms exhibit a highly skewed Zipfian distribution in any large text \citep{lignos2018morphology}. 
Thus, uniform sampling creates an unnatural bias towards low-frequency types.
Since high frequency is correlated with irregularity across many but not all languages \citep{bybee1991natural,fratini2014frequency,wu-etal-2019-morphological}, this creates a bias towards more regular and reliable training items.

We provide two alternatives for producing realistic or challenging data sets: \textbf{(1)} a frequency-weighted sampling strategy to achieve a more realistic distribution of out-of-vocabulary (OOV) lemmas and inflectional categories and better match practical use-cases or input during child language acquisition, and \textbf{(2)} a sampling strategy that explicitly balances OOV lemmas and inflectional categories in order to directly evaluate models' generalization ability along these dimensions.


\subsection{Practice 2: Single Data Splits} 

The current practice in inflection evaluation, employed, for example, in the SIGMORPHON, CoNLL-SIGMORPHON and SIGMORPHON-UniMorph shared tasks in recent years \citep{cotterell-etal-2016-sigmorphon,cotterell-etal-2017-conll,cotterell-etal-2018-conll,mccarthy-etal-2019-sigmorphon,vylomova-etal-2020-sigmorphon,pimentel-ryskina-etal-2021-sigmorphon,kodner-etal-2022-sigmorphon}, examines different models with one particular data set that is considered representative of the language or the inflection task at hand. This data set, and therefore all evaluation, usually consists of one pre-defined train-(dev-)test split.

However, this method is problematic because it implicitly assumes that the results from a single split are informative and generalizable. In reality, this assumption is untenable, particularly when facing severe data limitation~\citep{liu-prudhommeaux-2022-data}, as is the case for the majority of languages in the world (cf.~\citealp{blasi-etal-2022-systematic}):  In UniMorph 4, for example, data set size varies significantly across languages, with the smallest, Manx (Celtic, IE), containing only one lemma with 14 inflected forms, and the largest, Czech (Slavic, IE) containing approximately 800,000 lemmas with 50.3 million forms. If the performance on a single split is not necessarily representative, then the original model ranking derived from the one particular data split might also not generalize well.


The concerns outlined above were demonstrated in \citet{liu-prudhommeaux-2022-data}, which investigated model generalizability in low-resource morphological segmentation. Using data from 11 languages, they provided evidence that: \textbf{(1)} there are major differences in the numerical performance and rankings of each evaluated model type when using different splits from the same data set, and \textbf{(2)} even within a single split,  large performance variability can arise for each model type when it is trained using different random seeds. These findings illustrate that common methods of model evaluation can lead to largely coincidental conclusions.
We extend this approach to morphological inflection by applying multiple data splits, and evaluating variability between splits.

\subsection{Practice 3: Uncontrolled Overlaps} 

The typical morphological inflection task paradigm presents (\texttt{lemma}, \texttt{inflected form}, \texttt{feature set}) triples during training and asks a system to predict inflected forms from (\texttt{lemma}, \texttt{feature set}) pairs during evaluation. Note that since the lemma and feature set can be combined independently, it is possible for either lemmas or feature sets that appeared during training to reappear during test without any individual triple violating train-on-test. Test pairs with OOV lemmas or feature sets require a system to generalize along different morphological dimensions. Performance is likely related to the relative rates of OOV lemmas and feature sets in the evaluation split, yet existing sampling strategies generally leave these variables uncontrolled.

We observe that uncontrolled OOV rates vary dramatically between different sampled data splits, and that uncontrolled sampling biases test sets  towards ``easier'' items with in-vocabulary lemmas and feature sets.
To remedy this, we argue that performance should be reported independently for items with each lemma/feature set overlap type regardless of sampling strategy. Furthermore, if a project's research goal is to evaluate the generalization ability of a model, lemma/feature set overlap-aware sampling should be used to ensure that a sufficient number of test items of each overlap type are present.

\section{Defining Overlap}\label{sec:overlap}

Morphological inflection requires generalization over two primary dimensions: to new lemmas (``\textit{If I have witnessed the 2pl imperfective subjunctive with other verbs, how do I apply that to new verb X?}'') and to new inflectional categories (``\textit{If I have seen X inflected in several other categories, how do I create the 2pl imperfect subjunctive of X?}''). Because of the sparsity of morphological inflections in language use \citep{chan2008structures}, both types of generalization are necessary during language acquisition as well as deployment of computational models. 

As with many linguistic phenomena, the attestation of inflected forms follows an extremely sparse and skewed long-tailed distribution, as do attested lemmas ranked by the proportions of their potential paradigms that are actually attested (\textit{paradigm saturation}; PS), and inflectional categories 
ranked by the number of lemmas with which they occur \citep{chan2008structures}. For example, the median PS for Spanish verbs in millions of tokens of child-directed speech is equivalent to \textit{two} of its three dozen possible forms, and the 2nd person plural imperfect subjunctive only occurs with two lemmas \citep[cf.][]{lignos2018morphology,kodner2022computational}.

Given the importance of both types of generalization, it is necessary to evaluate both to assess the abilities of a morphological learning model.
In the evaluation made popular by the SIGMORPHON shared tasks, models are
asked to predict inflected forms given (\texttt{lemma}, \texttt{feature set}) pairs, where feature sets can be seen as corresponding to inflectional categories or paradigm cells. 
Generalization across lemmas is required when an evaluation pair contains a lemma that was out-of-vocabulary (OOV) in training, and generalization across categories is required when an evaluation pair contains a feature set that was OOV. In all, there are four logically possible licit types of evaluation pairs distinguished by their lemma and feature overlap with training. These are expressed visually in Figure \ref{fig:overlap_diagram} along with two types which are unions of the other types:
\begin{description}[noitemsep]
    \item[\both\ Overlap:] Both the lemma and feature set of an evaluation pair are attested in the training set (but not together in the same triple).
    \item[\lemma\ Overlap:] An eval pair's lemma is attested in training, but its feature set is novel.
    \item[\feats\ Overlap:] An eval pair's feature set is attested in training, but its lemma is novel.
    \item[\neither\ Overlap:] An evaluation pair is entirely unattested in training. Both its lemma and features are novel.\smallskip
    \item[\featsAttested:] An eval pair's feature set is attested in training (\both\ $\cup$ \feats)
    \item[\featsNovel:] An eval pair's feature set is novel (\lemma\ $\cup$ \neither)
\end{description}

For a concrete illustration, consider the training and evaluation sets provided in (\ref{exe:partition1})-(\ref{exe:partition2}). Each evaluation pair exhibits a different kind of overlap.

\begin{exe}
    \ex \textbf{Example Training Set}\label{exe:partition1}
    \vspace{-0.1cm}
    {\small\begin{verbatim}
t0: see  seeing  V;V.PTCP;PRS
t1: sit  sat     V;PST\end{verbatim}}
    \ex \textbf{Example Evaluation Set}\label{exe:partition2}
    \vspace{-0.1cm}
    {\small\begin{verbatim}
e0: see  V;PST       <-- both
e1: sit  V;NFIN      <-- lemmaOnly
e2: eat  V;PST       <-- featsOnly
e3: run  V;PRS;3;SG  <-- neither\end{verbatim}
\begin{verbatim}
featsAttested = {e0, e2}
featsNovel    = {e1, e3}\end{verbatim}}
\end{exe}

Computational work in morphological inflection has generally ignored these dimensions of evaluation. In the shared task, the four overlap types were uncontrolled before 2021, which contains one partial evaluation on \feats\ $\cup$ \neither\ items. But, recognition of the value of these overlap types has grown recently. \citet{goldman-etal-2022-un} showed that four models consistently struggle to generalize across lemmas, concluding that test sets should avoid lemma overlap altogether. However, this proposal removes the option to contrast performance on seen and unseen lemmas. Furthermore, they did not control for or evaluate feature overlap, so \both\ vs. \lemma\ and \feats\ vs. \neither\ also cannot be distinguished. 
(\ref{exe:partition_goldman}) summarizes their partition scheme, which distinguishes two overlap types. We call these \texttt{lemmaAttested} (= \both\ $\cup$ \lemma) and \texttt{lemmaNovel} (= \feats\ $\cup$ \neither).

\begin{exe}
    \ex \textbf{\citet{goldman-etal-2022-un} Partition Types\label{exe:partition_goldman}}
    \vspace{-0.1cm}
    {\small\begin{verbatim}
e0: sit  V;PST       <-- lemmaAttested
e1: see  V;NFIN      <-- lemmaAttested
e2: eat  V;PST       <-- lemmaNovel
e3: run  V;PRS;3;SG  <-- lemmaNovel\end{verbatim}}
\end{exe}

The 2022 SIGMORPHON-UniMorph shared task 
 was the first to report results on all four overlap types (\both, \feats, \lemma, \neither). Every system submitted to the shared task achieved much better performance on in-vocabulary feature sets (\both\ and \feats) than OOV feature sets (\lemma\ or \neither). This discrepancy even held for languages for which a model should be able to generalize:  highly regular agglutinative morphology for which this type of generalization is often transparent. On the other hand, lemma attestation produced a much smaller discrepancy. 
 Following these observations, we focus our investigation on the four logical overlap types with extra emphasis on the \featsAttested\ vs. \featsNovel\ dichotomy. We address agglutinative languages specifically in Section \ref{sec:overlap_analysis}

\section{Data Sources and Preparation}\label{sec:data}

We follow prior literature in providing training and evaluation data in UniMorph's format. Data sets were sampled from UniMorph 4 \citep{batsuren-etal-2022-unimorph} and 3 \citep{mccarthy-etal-2020-unimorph}\footnote{In some cases, UniMorph 4 was found to lack high-frequency items present in UniMorph 3. For example, English verbs \textit{happen} and \textit{run} are present in 3 and absent in 4. For languages where we determined this to be an issue, we sampled from deduplicated UniMorph 3+4 with tags normalized to 4.} augmented with frequencies from running text corpora. When possible, frequencies were drawn from child-directed speech (CDS) corpora from the CHILDES database \citep{macwhinney2000childes}, since one possible downstream application of the morphological inflection task is contribution to  the computational cognitive science of language acquisition. CHILDES lemma and morphological annotations were converted into UniMorph format and intersected with UniMorph to create frequency lists.\footnote{All data and code is available at \url{https://github.com/jkodner05/ACL2023_RealityCheck}.}

\subsection{Languages}

Languages were prioritized for typological diversity and accessibility of text corpora. Quantitative summaries of our frequency+UniMorph data sets are provided in Appendix \ref{sec:appendix-data}.

\textbf{Arabic (Semitic, AA):} Modern Standard Arabic frequencies were drawn from the diacritized and morphologically annotated Penn Arabic Treebank \citep[PATB;][]{maamouri2004penn} and intersected with UniMorph 4 \texttt{ara} $\cup$ \texttt{ara\_new}. Diacritized text is a requirement because orthographic forms drawn from undiacritized text are massively morphologically ambiguous. The text in the CHILDES Arabic corpora is undiacritized and thus unusable. 

\textbf{German (Germanic, IE):} German was drawn from the Leo Corpus \citep{behrens2006input}, the only morphologically annotated German corpus in CHILDES, and intersected with UniMorph 3+4. Only nouns and verbs were extracted because annotation for adjectives is inconsistent. 

\textbf{English (Germanic, IE):} English was included because it is heavily studied despite its relatively sparse morphology. Data was extracted from all morphologically annotated CHILDES English-NA corpora and intersected with UniMorph 3+4.\footnote{A full list of utilized English and Spanish CHILDES corpora is provided in Appendix \ref{sec:appendix-childes}.} Only nouns and verbs were extracted due to inconsistent adjective annotation in both data sources. 

\textbf{Spanish (Romance, IE):} Spanish exhibits a variety of fusional and agglutinative patterns. Data was extracted from all morphologically annotated Spanish CHILDES 
corpora intersected with Spanish UniMorph 3+4. Non-Spanish vocabulary was removed by intersecting with UniMorph. Only nouns and verbs were extracted.

\textbf{Swahili (Bantu, Niger-Congo):} Swahili morphology is highly regular and agglutinative with very large paradigms. Frequencies were drawn from Swahili Wikipedia dump 20221201 accessed through Huggingface \citep{wikidump} and intersected with UniMorph 4 \texttt{swc} $\cup$ \texttt{swc.sm}. In cases where mapping inflected forms to UniMorph creates ambiguity due to syncretism, frequency was divided evenly across each triple sharing the inflected form. This ensured that the frequencies of inflected forms remain consistent with Wikipedia. Intersecting with UniMorph removed the large amount of non-Swahili vocabulary in the Wikipedia text.

\textbf{Turkish (Turkic):} Turkish is also highly regular and agglutinative with very large paradigms. Frequencies were drawn from Turkish Wikipedia dump 20221201 accessed through Huggingface, intersected with UniMorph 4, and processed identically to Swahili. 

\subsection{Data Splits}

We employed three distinct sampling strategies to generate small (400 items) and large (1600) training, small (100) and large (400) fine-tuning, development (500), and test (1000) sets for each language.\footnote{Swahili large train and large fine-tune contain 800 and 200 items respectively due to the limited size of UniMorph.} Small training and fine-tuning are subsets of large training and fine-tuning. Each splitting strategy was applied five times with unique random seeds to produce distinct data sets.

\textbf{\uniform:} Raw UniMorph 3+4 corpora were partitioned uniformly at random. This approach is most similar to that employed by SIGMORPHON shared tasks, except for 2017 and 2022.

\textbf{\weighted:} 
Identical to \uniform\ except splits were partitioned at random weighted by frequency. Small training+fine-tuning were sampled first, then additional items were sampled to create large training+fine-tuning. Training and fine-tuning sets were then split uniformly at random. Dev+test was next sampled by weight and then were separated uniformly. This frequency-weighted sampling is reminiscent of the 2017 shared task: it strongly biases the small training set towards high-frequency items and dev+test towards low-frequency items. Since most UniMorph forms do not occur in our corpora due to morphological sparsity, most triples had zero weight and were never sampled. 

\textbf{\aware:} Similar to the 2022 SIGMORPHON shared task. It enforces a maximum proportion of \featsAttested\ pairs in the test set relative to train+fine-tuning: as close to 50\% as possible without exceeding it. This ensures that there is ample representation of each overlap type in test.
It is  adversarial, since \featsNovel\  pairs are expected to be more challenging than \featsAttested\ pairs. This process also tends to increase the proportion of \texttt{lemmaOnly} items in the test set. Only items with non-zero frequency were sampled. 
\smallskip

\uniform\ produces a heavy bias towards lower frequency words. For all languages and splits, the median frequency of sampled items is actually zero: that is, the majority of sampled items were not attested in our corpora. This is a consequence of the extreme sparsity of morphological forms discussed in Section \ref{sec:overlap}. As a consequence, overlap between splits from different seeds is orders of magnitude lower for \uniform\ than the other strategies. \weighted\ achieves the expected high-frequency bias in training sets relative to test sets. 

Table \ref{tab:strategy-summary2} provides average means and standard deviations for the proportion of \featsAttested\ and \featsNovel\ in test sets relative to small and large train. \aware\ consistently achieves a roughly 50-50 split with low variability across languages and seeds. The other strategies bias test sets heavily towards \featsAttested\ with high variance across languages and seeds.\footnote{See Appendix \ref{sec:appendix-data} for breakdowns by language, training size, and overlap partitions.}

\begin{table}[h]
    \centering
    \small
    \begin{tabular}{l|ccc}
    \textbf{Test vs S Train} & $\mu$ \%\featsAttested  & $\sigma$\\
    \hline
    \uniform &  80.33\%  & 19.50\% \\
    \weighted & 90.44    & 11.13\\
    \aware &  48.81  & 0.98\\
    \hline
    \hline
    \textbf{Test vs L Train} &  $\mu$ \%\featsAttested  & $\sigma$\\
    \hline
    \uniform &  96.17\%  & 5.55\% \\
    \weighted &  95.36  & 7.28\\
    \aware &  49.92   & 0.17\\
    \end{tabular}
    \caption{Language-by-language average mean percentage and standard deviation for proportion of \featsAttested\ attested in test relative to small and large training. \%\featsNovel = 100 - \%\featsAttested.}\vspace{-0.3cm}
    \label{tab:strategy-summary2}
\end{table}

\section{Experimental Setup}\label{sec:experiment}

One non-neural and three neural systems were evaluated. These were chosen based on their availability and performance in recent  shared tasks:

\textbf{\wuetal} \citep{wu-etal-2021-applying}  is a character-level transformer that was used as a baseline in 2021 and 2022. We used the hyper-parameters suggested by the original authors for small training conditions.

\textbf{\cluzhgr} and \textbf{\cluzhbfour} \citep{wehrli-etal-2022-cluzh} is a character-level transducer which substantially outperformed \wuetal\ in the 2022 shared task. The results submitted for the shared task are from an elaborate ensemble model optimized for each language. For this work, we evaluate two published variants with consistent hyper-parameters across languages, \cluzhgr\ with greedy decoding and \cluzhbfour\ with beam decoding, beam size = 4.

\textbf{\nonneural} \citep{cotterell-etal-2017-conll} has been used as a baseline in SIGMORPHON shared tasks since 2017. It heuristically extracts transformations between lemmas and inflected forms and applies a majority classifier conditioned on the associated feature sets. \nonneural\ was trained on combined training and fine-tuning sets so that each architecture was exposed to the same amount of data.


\section{Results}\label{sec:results}

This section presents our analyses of the results. All evaluations report exact match accuracy. \textit{Overall accuracy} refers to average accuracy on an entire evaluation set. \textit{Average overall accuracy} refers to the mean of overall accuracy over all five seeds. 
See Appendix \ref{sec:appendix-results} for 
full breakdowns by language and architecture.

\subsection{Effect of Training Size}

We begin by comparing average overall accuracy for each training size. All reported analyses focus on test, but there were no observable qualitative differences in behavior between dev and test. We summarize the results in Table \ref{tab:accuracy-summary}, broken down by overlap partition and sampling strategy.  The large training size consistently leads to higher accuracies than small training. Across languages, the average accuracy score difference between the two training sizes is 9.52\%. Taking Arabic as an illustrative example, the score difference between the two training sizes ranges from 1.74\% to 19.32\% depending on model type and splitting strategy, with an average of 12.05\%.

\begin{table}[h]
    \centering
    \small
    \begin{tabular}{l|cc}
    \textbf{Test vs S Train} & \featsAttested & \featsNovel\\
    \hline
\uniform & 70.47\% & 33.57\% \\
\weighted & 79.25 & 22.77 \\
\aware & 79.60 & 31.13 \\
    \hline
    \hline
    \textbf{Test vs L Train} &  \featsAttested & \featsNovel\\
    \hline
\uniform & 80.00\% & 55.57\%  \\
\weighted & 85.94 & 23.74  \\
\aware & 86.22 & 35.51  \\
    \end{tabular}
    \caption{Overall accuracy across languages by overlap type in test.}\vspace{-0.3cm}
    \label{tab:accuracy-summary}
\end{table}


\begin{figure*}[t]
    \includegraphics[width=\textwidth]{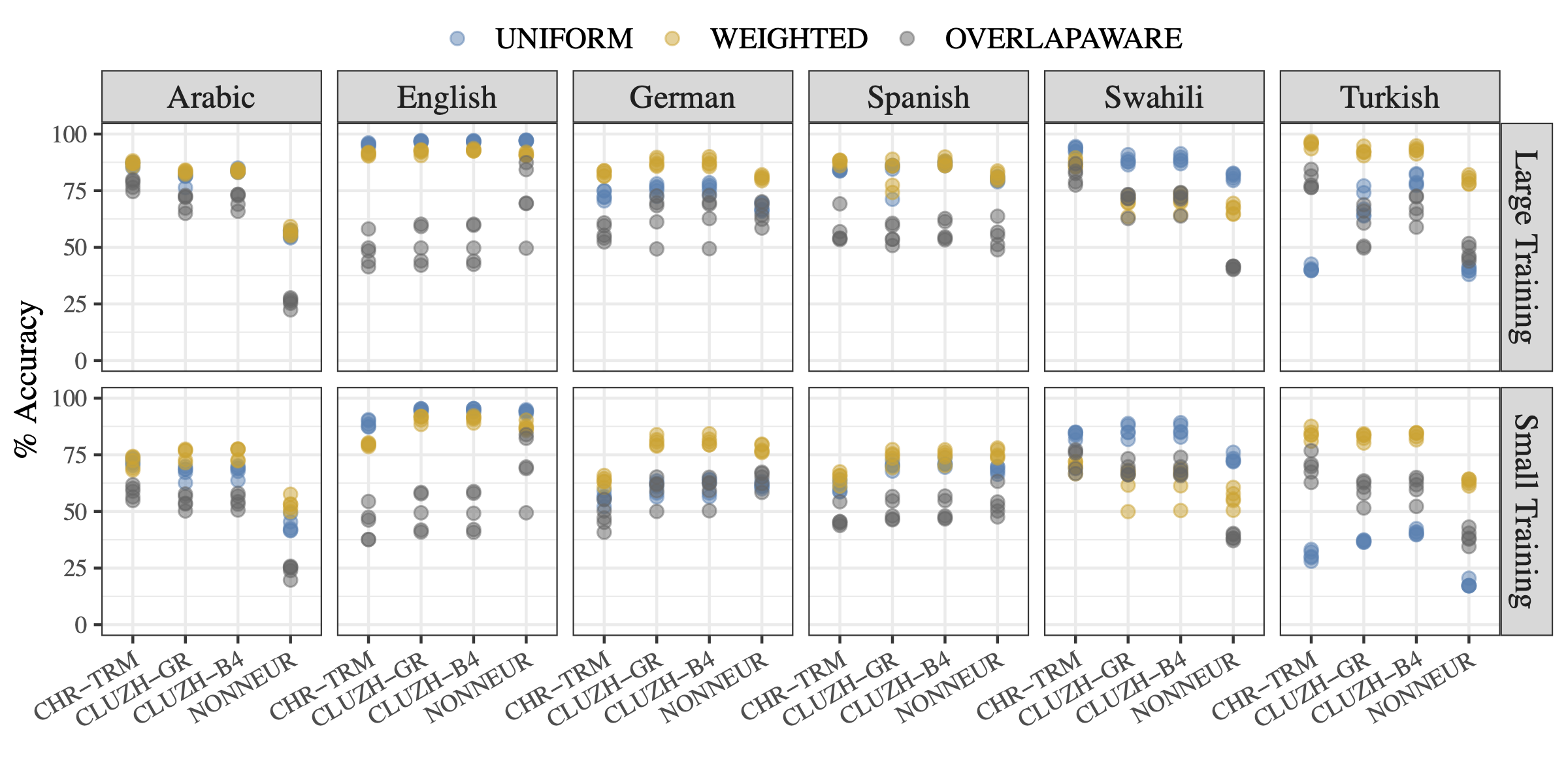}
        \vspace{-1.2cm}
    \caption{Overall accuracy for each language/seed by training size, sampling strategy, and model type.}
            \vspace{-0.3cm}
    \label{fig:splittype}
\end{figure*}

\subsection{Effect of Sampling Strategy}


We next turn to measuring the effect of sampling strategy on overall accuracy. Figure \ref{fig:splittype} provides a visualization of accuracy by sampling strategy across seeds broken down by training size, language, model type. Using Arabic as an illustration, for large training, {\weighted} sampling leads to the highest average overall accuracy across model types (77.76\%), while {\aware} sampling yields the lowest (61.06\%); comparing the results from the three sampling strategies given each of the four model types, {\weighted} consistently results in the highest accuracy  for all model types except for {\cluzhbfour}, where {\uniform} sampling (83.84\%) leads to a performance slightly better than that of {\weighted} (83.82\%). 
We make similar observations for small training: {\weighted} and {\aware} result in the highest and the lowest average overall accuracy, respectively, across model types  for Arabic (68.82\% vs. 47.81\%). {\weighted} sampling leads to a higher accuracy compared to the other two strategies for every model type other than {\wuetal}, 
where the result from {\uniform} sampling (71.90\%) is again slightly higher than that of {\weighted} (71.60\%).

When considering other languages, we also find some variation. {\weighted} sampling also yields the highest average accuracy scores across model types for Arabic, German, Spanish, and Turkish for both training sizes, except for Spanish under the large training condition with {\cluzhgr}, where {\uniform} leads. 
In contrast, {\uniform} consistently results in the highest average accuracy on English and Swahili for both training sizes. 

Across languages, the average accuracy from {\weighted} is the highest for both large (83.75\%) and small (74.22\%) training sizes, followed by {\uniform} (large: 79.20\%, small: 66.16\%). {\aware} always yields the lowest accuracy. These observations align with our expectations about the adversarial nature of {\aware}, where challenging \featsNovel\ (Table \ref{tab:accuracy-summary}) constitutes a much larger proportion test set (Table \ref{tab:strategy-summary2}).

\subsection{Effect of Overlap}
\label{sec:overlap_analysis}

\begin{figure*}[t]
    \includegraphics[width=\textwidth]{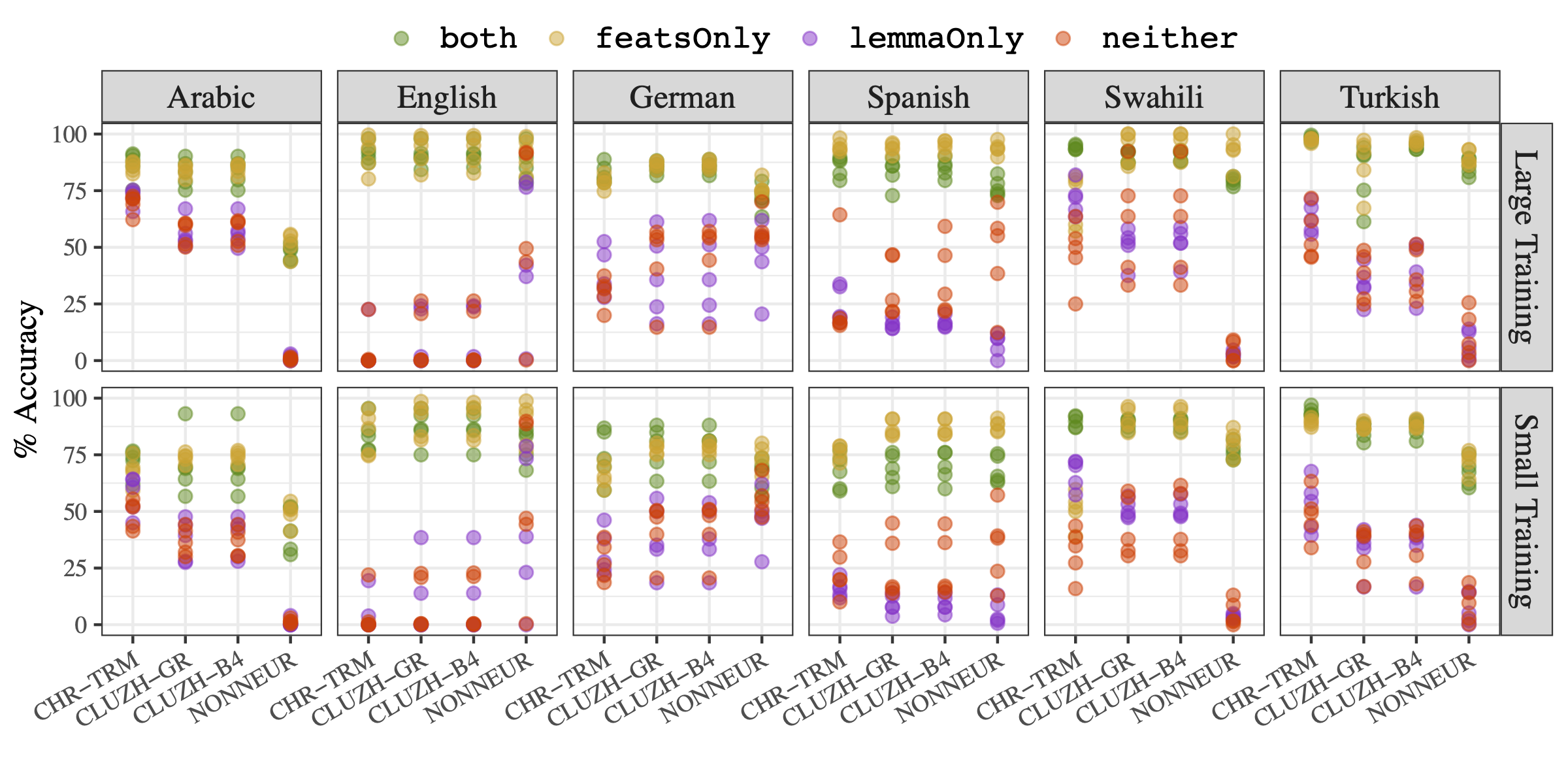}
        \vspace{-1.2cm}
    \caption{Accuracy on \aware\ splits for each partition/seed by training size, language, and model type. \featsAttested\ = \both\ (green) and \feats\ (gold). \featsNovel\ = \lemma\ (violet) and \neither\ (red).}
                \vspace{-0.3cm}
    \label{fig:overlaptype}
\end{figure*}

We now provide an analysis of accuracy scores  by overlap partition. Figure \ref{fig:overlaptype} provides a visualization of accuracy by partition across seeds broken down by training size, language, model type. Using Arabic again as an illustration, the average accuracy across model types and sampling strategies for large training is much higher for {\featsAttested} (77.70\%) than for {\featsNovel} (41.92\%),   somewhat higher accuracy is achieved for  {\both} (79.53\%) than for {\feats} (77.28\%), and higher accuracy is achieved for {\lemma} (49.12\%) than for {\neither} (41.92\%). 
This ranking is consistent across model types, sampling strategies, and training sizes. 
Scores from these two overlap partitions are also higher than those from {\lemma} and {\neither}. 

These patterns hold across languages. Specifically, we observe two general tendencies. First,  the accuracy averaged across model types and sampling strategies is always substantially higher for {\featsAttested} than it is for {\featsNovel}; the average accuracy difference between the two is 49.75\% for the large training, and 48.02\% for small training. This is reflected in a full breakdown by overlap type: higher accuracy is consistently achieved for {\both} and {\feats}, than for {\neither} and {\lemma}. 
This large asymmetry corresponds to our expectations regarding the effect of feature overlap on performance.

We provide three sub-analyses to further investigate this asymmetry and compare it with the lemma-based division advocated for by \cite{goldman-etal-2022-un}. First, we compute the average accuracy difference between \texttt{lemmaAttested} ({\both} $\cup$ {\lemma}) and \texttt{lemmaNovel} ({\feats} $\cup$ {\neither}). The score difference between \texttt{lemmaAttested} and \texttt{lemmaNovel} is less than 2\% averaged across languages for both training sizes, which is an order of magnitude smaller than the difference between {\featsAttested} and {\featsNovel}. This trend is consistent with the results of the 2022 SIGMORPHON shared task, which also found a much greater impact of feature set attestation than lemma attestation.

Second, we measure the correlation between the proportion of {\featsAttested} items (number {\featsAttested} items divided by the size of the dev or test set), and overall accuracy (average accuracy on an entire dev or test set), as well as  between the proportion of \texttt{lemmaAttested} and overall accuracy. We used Spearman's $\rho$, which assesses if there is any monotonic (not necessarily linear) relationship between the two variables.\footnote{$\rho$ falls in the range [-1,1], where -1 is a perfect negative correlation and 1 is a perfect positive correlation.} If $\rho$ between an overlap type and overall accuracy is high, it would suggest that the distribution of overlaps is an important driver of performance. \texttt{lemmaAttested} shows little correlation (small: 0.01, large: -0.10). However, we find substantial positive correlations for {\featsAttested} (small: 0.69, large: 0.68).


Third, we compute the correlation between the accuracy score of individual partitions and the overall accuracy score on \uniform\ and \weighted\ vs. on \aware. This demonstrates to what extent evaluation results based on each overlap partition resemble those captured by the overall accuracy and how it differs when overlaps are controlled during sampling. If the correlation is small, it suggests that the performance on a particular overlap partition is largely independent of the others and should be evaluated independently. 

When overlaps are not explicitly controlled, correlations are particularly strong for \featsAttested\ because this partition makes up a large majority of the test set (Table~\ref{tab:metric_corr}). These partitions are also the ones that tend to show the highest performance, which is then reflected in the overall accuracy.
However, for \aware, correlations are higher between overall accuracy and the challenging partitions: \featsNovel, \lemma, and \neither. They are also higher  not only for \featsNovel, but also \texttt{lemmaAttested}, and \texttt{lemmaNovel} even though these overlaps were not explicitly controlled. This demonstrates that \aware\ sampling better balances individual partitions in its overall accuracy scores and can be expected to produce a more challenging evaluation. However, all partitions should be evaluated regardless of sampling strategy.

\begin{table}[h]
\footnotesize
    \centering
    \begin{tabular}{l|c|c}
      \textbf{Overlap Partition}   & \textbf{Uncontrolled $\rho$} & \textbf{Controlled  $\rho$}  \\\hline\hline
     {\featsAttested}   & 0.97 & 0.45\\
     {\featsNovel} & 0.16 &  0.93\\
     \hline
     \texttt{lemmaAttested} & 0.84 & 0.88\\
     \texttt{lemmaNovel} & 0.78 & 0.82\\
     \hline
     {\both} & 0.89 &  0.49\\
     {\feats} & 0.73 & 0.21 \\
     {\lemma} & 0.24 & 0.89 \\
     {\neither} & -0.04 & 0.85\\
    \end{tabular}
    \caption{Correlation between average accuracy for each overlap partition and average overall accuracy across the six languages. Uncontrolled = \weighted\ and \uniform. Controlled = \aware.}\vspace{-0.3cm}
    \label{tab:metric_corr}
\end{table}

Up to this point, we have considered all languages in the analysis. However, whether or not it is reasonable to expect a system to achieve high accuracy on \featsNovel\ items varies typologically. For languages with highly regular and agglutinative morphologies, such as Swahili and Turkish, each feature in a feature set roughly corresponds to a single affix in a certain order with a limited number of allomorphs. For these languages, this dimension of generalization should often be straightforward. For languages with mixed systems, like Spanish and Arabic, and languages with fusional systems like English, the individual members of a feature set often do not have direct bearing on the inflected form. For these languages, generalization to a novel feature set is sometimes impossible when it cannot be inferred from its component features. The same problem applies to lemmas with erratic stem changes or suppletion.

Thus, if a model type can generalize to novel feature sets, one would expect that the accuracy gap between {\featsAttested} and {\featsNovel} would be lower for Swahili and Turkish than for the other languages. However, the gaps for these are actually larger than for German or Arabic. One would also expect the correlation between the proportion of \featsAttested\ in the data and overall accuracy to be lower for Swahili and Turkish, however this is not borne out either. These findings, provided in Table \ref{tab:turkishswahili}, reveal that current leading inflection models do not necessarily generalize well to novel feature sets even in precisely the cases where they should be able to.

\begin{table}[h]
\footnotesize
    \centering
    \begin{tabular}{l|c|c|c}
    \textbf{Train}     &  \textbf{Language} & \textbf{Avg. Score} & \textbf{\featsAttested}\\
     \textbf{Size}   &  \textbf{Strategy} & \textbf{Difference} & \textbf{$\sim$Accuracy $\rho$}\\\hline\hline
   Small & {Arabic}  & 33.00\% & 0.57 \\
       &   {Swahili} & 40.04 & 0.63 \\
    &   {German} & 40.35 & 0.23 \\
    &   {Turkish} & 41.96 & 0.83 \\
    &   {Spanish} & 52.60 & 0.75 \\
   & {English}  & 74.10 & 0.66 \\
   \hline
   Large & {Arabic}  & 35.79\% & 0.44 \\
      &   {German} & 36.19 & 0.73 \\
    &   {Swahili} & 39.26 & 0.64 \\
    &   {Turkish} & 52.14 & 0.59 \\
    &   {Spanish} & 61.01 & 0.64 \\
   & {English}  & 80.17 & 0.82 \\
    \end{tabular}
    \caption{Avg. score difference between \featsAttested\ and \featsNovel\ and correlation between proportion \featsAttested\ and overall accuracy by language/training size, ranked by score difference. }\vspace{-0.3cm}
    \label{tab:turkishswahili}
\end{table}




\subsection{Model Ranking}
\label{ranking}


In this section, we analyze how performance varies across the four model types. We first compare model performance based on the average overall accuracy. Averaged across the six languages, {\cluzhbfour} ranks among the highest, while {\nonneural} consistently achieves the lowest performance.

{\footnotesize
\begin{description}[noitemsep]
    \item[large:] {\cluzhbfour} (78.32\%) $>$ {\wuetal} (78.07\%) $>$ {\cluzhgr} (76.17\%) $>$ {\nonneural} (65.82\%)
    \item[small:] {\cluzhbfour} (68.58\%) $>$ {\cluzhgr} (67.97\%) $>$ {\wuetal} (64.76\%) $>$ {\nonneural} (58.97\%)
\end{description}
}

Model rankings for individual languages are much more variable, especially for large training. There is not a single model ranking that holds for every language. While {\cluzhbfour} yields the best performance for three languages (German, Spanish, and Turkish), {\wuetal} outperforms other model types for Arabic and Swahili, and {\nonneural} leads to the highest accuracy for English. 
There is less variation in model rankings for small training; the same model ranking was observed for German, English, and Spanish ({\nonneural} $>$ {\cluzhbfour} $>$ {\cluzhgr} $>$ {\wuetal}). Notably, for each individual language, the model rankings were always inconsistent between the two training sizes. 

Several trends emerge in model rankings by overlap partition. First, the model rankings based on the overall accuracy do not hold for the overlap partitions except for Arabic and Swahili large training. Second, within each overlap partition, model rankings are more stable across languages for small train than large. Third, on average, {\cluzhbfour} outperforms the other model types on partitions with feature overlap whereas {\wuetal} leads on partitions without feature overlap. These tendencies resonate with our proposal in Section~\ref{sec:overlap}: future models of morphological inflection should be evaluated based on alternative metrics in addition to overall accuracy. They also reveal difference generalization strengths across models.

When comparing performance by sampling strategy, we found lower variability for each language. For example, with  {\uniform} large training, two model rankings turn out to be the most frequent, each observed in two languages. Among the models, {\cluzhbfour} and {\wuetal} achieve the best performance. For small training, one model ranking holds for three out of the six languages ({\cluzhbfour} $>$ {\cluzhgr} $>$ {\wuetal} $>$ {\nonneural}). Considering both training sizes, there are no noticeable differences in terms of the most frequent model ranking across the three sampling strategies. For  {\uniform}  and  {\weighted}, the neural systems are always ranked among the highest for both training sizes; yet for {\aware} with small training, {\nonneural} achieves the highest performance for German, English, and Spanish.

\subsection{Variability across Random Seeds}

Analysis so far relies on accuracy scores averaged across random seeds. The final component of our analysis investigates how much variation arises due to random data sampling. Given the five random seeds for each combination of  language, sampling strategy, overlap partition, and model type, we calculated the \textit{score range}, which is the difference between the lowest and the highest overall accuracy, as well as the standard deviation of the accuracy scores across the seeds, which we refer to as \textit{random seed variability}.

We first considered the score range for overall accuracy for each language. For large training, the mean score range spans from 4.41\% for Arabic, to 8.38\% for English; the mean random seed variability follows the same trend (1.73\% to 3.54\%). For every language, the score range and random seed variability for the large training size are consistently larger than those derived from small training. In both cases, score ranges are non-negligible. 

\begin{table}[h]
\footnotesize
    \centering
    \begin{tabular}{l|c|c|c}
    \textbf{Train}     &  \textbf{Sampling} & \textbf{Score} & \textbf{Random Seed}\\
     \textbf{Size}   &  \textbf{Strategy} & \textbf{Range} & \textbf{Variability}\\\hline\hline
   Small & {\uniform}  & 4.51\% & 1.84\% \\
   & {\weighted}  & 6.33 & 2.57 \\
   &   {\aware} & 12.13 & 5.01 \\\hline
   Large & {\uniform} & 3.99\% & 1.68\% \\
      & {\weighted}  & 4.08 & 1.66 \\
  & {\aware}  & 13.06 & 5.50 \\

    \end{tabular}
    \caption{Average score range and random seed variability across languages for each sampling strategy for both training sizes.}
    \label{tab:seeds}
\end{table}

Next, for each language, we analyze the average score range for each sampling strategy and model type separately. Comparing results from the three sampling strategies in Table~\ref{tab:seeds}, {\aware} sampling consistently yields the highest score range and random seed variability. 
This indicates that {\aware}, despite  exhibiting the least variability in overlap partition sizes, is also the most variable in terms of model performance. This likely suggests that it is not just feature set attestation in general, but also exactly which feature sets that happen to appear in train vs. test drive performance.
Finally, when looking at results for each individual model type, {\cluzhgr} demonstrates the most variable performance. Its average score range (9.47\% for large training, 7.94\% for small) and its average random seed variability (4.03\% for large training, 3.31\% for small) end up being the highest.

\section{Conclusions}\label{sec:discussion}

We investigated the roles that sampling strategy, random seeds, and overlap types play in evaluating and analyzing the results of morphological inflection tasks and conclude that common practices leave much to be desired. We argue for frequency-weighted splitting to achieve more realistic train-test distributions and feature/lemma overlap-aware sampling for directly investigating the generalization abilities of different models. The high score range observed for overlap-aware sampling relative to other strategies suggests that which feature sets happen to appear in train vs. test play a major role in the ability of a model to generalize, though future work would need to confirm this. 

Regardless of sampling strategy, evaluation items of each overlap type should be used in addition to an overall analysis. The evaluation in this work reveals that all model types under investigation struggle to generalize to unseen feature sets, even for languages where that should be possible, a fact that has been overlooked in prior studies. Finally, results drawn from one data split are unlikely to be representative, so multiple splits should be made with different random seeds and compared, particularly for shared tasks and leader boards where final model rankings matter.

\section*{Limitations}

Our suggested approaches have two primary practical limitations: First, \weighted\ sampling is restricted to languages with available running text sources for extracting frequencies. A project on \textit{extremely} low-resource languages \cite[e.g.,][]{liu-etal-2022-always} may be restricted to \uniform\ and \aware\ sampling.
Second, as the number of seeds increases, so do requirements for training time and/or computing power. A shared task, for example, might limit itself to only a few seeds in order to assure on-time submissions. Future work would benefit from a wider selection of model architectures, along with more sampling strategies, and of course a wider sample of typologically diverse languages.

Notably, this work reproduces the effect observed in the SIGMORPHON 2022 shared task \citep{kodner-etal-2022-sigmorphon}, which found a substantial performance hit for \featsNovel\ relative to \featsAttested, but not \texttt{lemmaNovel} relative to \texttt{lemmaAttested}. However, both this work and the shared task fail to replicate the effect observed in \citet{goldman-etal-2022-un}, which reports a 95\% performance hit on \texttt{lemmaNovel} vs. \texttt{lemmaAttested}. This may have something to do with differences in splitting algorithms, unmeasured feature overlap in \citet{goldman-etal-2022-un}, or choice of model architectures.

\section*{Ethics Statement}
To the best of our knowledge, all results published in this paper are accurate, and we have represented prior work fairly to the best of our abilities. All data sources are free and publicly available, except for the Penn Arabic Treebank \citep{maamouri2004penn}, which is accessible through the LDC.\footnote{\url{https://catalog.ldc.upenn.edu/LDC2005T20}} No sensitive data was used which could violate individuals' privacy or confidentiality. Authorship and acknowledgements  fairly reflect contributions.

\section*{Acknowledgements}
We thank Charles Yang, Jeffrey Heinz, Mitch Marcus, and the audience at Stony Brook University ATLaC for their helpful discussion. Experiments were performed on the SeaWulf HPC cluster maintained by RCC and the Institute for Advanced Computational Science (IACS) at Stony Brook University and made possible by National Science Foundation (NSF) grant No. 1531492. The second author gratefully acknowledges funding through the IACS Graduate Research Fellowship and the NSF 
Graduate Research Fellowship Program under NSF Grant No. 2234683.

\bibliography{anthology,custom}
\bibliographystyle{acl_natbib}

\appendix

\section{English and Spanish Data Sources}
\label{sec:appendix-childes}

\subsection{English}

The following CHILDES corpora were used to create the English data set. Utterances from speaker \texttt{*CHI} were excluded: 
Bates  \citep{bates1991first}, 
Bliss \citep{bliss1988development}, 
Bloom \citep{bloom1970language,bloom1974imitation}, 
Bohannon \citep{bohannon1977children}, 
Braunwald \citep{braunwald1971mother}, 
Brent \citep{brent2001role}, 
Brown \citep{brown1973first}, 
Clark \citep{clark1978awareness}, 
Davis \citep{davis1995articulatory}, 
Demetras \citep{demetras1986working,demetras1989changes}, 
EllisWeismer \citep{heilmann2005utility}, 
Feldman \citep{feldman1998constructing}, 
Garvey \citep{garvey1973social}, 
Gathercole \citep{gathercole1986acquisition}, 
Gelman \citep{gelman1998beyond}, 
Gillam \citep{gillam2004tnl}, 
Gleason \citep{gleason1980acquisition}, 
Hall \citep{hall1979communicative}, 
Higginson \citep{higginson1985fixing}, 
HSLLD \citep{dickinson2001beginning}, 
Kuczaj \citep{kuczaj1977acquisition}, 
MacWhinney \citep{macwhinney1991childes}, 
McCune \citep{mccune1995normative}, 
Morisset \citep{morisset1995toddlers}, 
Nadig \citep{bang2015learning}, 
Nelson \citep{nelson2006narratives}, 
NewEngland \citep{ninio1994classifying}, 
NewmanRatner \citep{newman2016input}, 
Nichols-TD \citep{nicholas1997communication}, 
Peters \citep{peters1987role}, 
POLER \citep{berl2005seizure}, 
Post \citep{demetras1986feedback}, 
Providence \citep{demuth2006word}, 
Rollins \citep{rollins2003caregivers}, 
Sachs \citep{sachs1983talking}, 
Sawyer \citep{sawyer2013pretend}, 
Snow \citep{macwhinney1985child}, 
Soderstrom \citep{soderstrom2008acoustical}, 
Sprott \citep{sprott1992children}, 
Suppes \citep{suppes1974semantics}, 
Tardif \citep{macwhinney2000childes}, 
Valian \citep{valian1991syntactic}, 
VanHouten \citep{vanhouten1986role}, 
VanKleeck \citep{macwhinney2000childes}, 
Warren-Leubecker \citep{warren1982sex}, 
Weist  \citep{weist2009finiteness}.

\subsection{Spanish}

The following CHILDES corpora were used to create the Spanish data set. Utterances from speaker \texttt{*CHI} were excluded: 
Aguirre \citep{martinez2005como}, 
ColMex \citep{macwhinney2000childes}, 
Fernandez/Aguado \citep{macwhinney2000childes}, 
GRERLI \citep{macwhinney2000childes}, 
Hess \citep{zimmermann2003desarrollo}, 
Linaza \citep{linaza1981lenguaje}, 
Marrero \citep{capelli1994aplicacion}, 
Montes \citep{montes1987secuencias}, 
AguadoOrea/Pine \citep{aguado2015comparing}, 
Ornat \citep{lopez1997lies}, 
Remedi \citep{remedi2014creacion}, 
SerraSole  \citep{macwhinney2000childes}.

\section{Splitting Strategy Data Summaries}
\label{sec:appendix-data}

This appendix contains Tables \ref{tab:strategy-summary1_bylang}-\ref{tab:strategy-summary2-bylang}.

\begin{table}[h!]
    \centering
    \scriptsize
    \begin{tabular}{l|cc|cc}
    \multicolumn{1}{c}{}  & \multicolumn{2}{c}{\textbf{Train}} & \multicolumn{2}{c}{\textbf{Test}} \\
    \textbf{Arabic} & $\mu\mu$ & $\mu M$ & $\mu\mu$ & $\mu M$\\
    \hline
    \uniform &  0.46 & 0  &  0.47 & 0  \\
    \weighted &  57.53 & 18  &  26.44 & 12  \\
    \aware &  6.72 & 2  &  6.46 & 2  \\
    \hline
    \hline
    \textbf{English} & $\mu\mu$ & $\mu M$ & $\mu\mu$ & $\mu M$\\
    \hline
    \uniform &  9.71 & 0  &  1.24 & 0  \\
    \weighted &  1840.51 & 362  &  122.55 & 67  \\
    \aware &  182.29 & 5  &  163.22 & 5  \\
    \hline
    \hline
    \textbf{German} & $\mu\mu$ & $\mu M$ & $\mu\mu$ & $\mu M$\\
    \hline
    \uniform &  0.14 & 0  &  0.18 & 0  \\
    \weighted &  111.99 & 20  &  9.56 & 5  \\
    \aware &  25.46 & 2  &  30.42 & 2  \\
    \hline
    \hline
    \textbf{Spanish} & $\mu\mu$ & $\mu M$ & $\mu\mu$ & $\mu M$\\
    \hline
    \uniform &  0.12 & 0  &  0.13 & 0  \\
    \weighted &  119.15 & 29  &  13.89 & 8  \\
    \aware &  25.50 & 2  &  21.97 & 2 \\
    \hline
    \hline
    \textbf{Swahili} & $\mu\mu$ & $\mu M$ & $\mu\mu$ & $\mu M$\\
    \hline
    \uniform &  40.13 & 0  &  38.38 & 0  \\
    \weighted &  518.95 & 88  &  8.11 & 4  \\
    \aware &  130.00 & 3  &  143.39 & 3  \\
    \hline
    \hline
    \textbf{Turkish} & $\mu\mu$ & $\mu M$ & $\mu\mu$ & $\mu M$\\
    \hline
    \uniform &  26.63 & 0  &  26.6 & 0  \\
    \weighted &  4854.13 & 1252  &  588.76 & 348  \\
    \aware &  436.41 & 12  &  397.94 & 12  \\
    \end{tabular}
    \caption{Average training and test item mean corpus frequency ($\mu\mu$) and median frequency ($\mu M$).}
    \label{tab:strategy-summary1_bylang}
\end{table}

\begin{table}[h!]
    \centering
    \scriptsize
    \begin{tabular}{l|c|c}
    \textbf{Arabic}  & $J_{LTrain}$ & $J_{Test}$ \\
    \hline
    \uniform & 0.10 & 0.05 \\
    \weighted & 9.90 & 3.17 \\
    \aware & 1.56 & 1.07\\
    \hline
    \hline
        \textbf{English}  & $J_{LTrain}$ & $J_{Test}$ \\
    \hline
    \uniform & 0.12 & 0.09 \\
    \weighted & 32.12 & 8.86\\
    \aware & 4.78 & 3.31\\
    \hline
    \hline
        \textbf{German}  & $J_{LTrain}$ & $J_{Test}$ \\
    \hline
    \uniform & 0.13 & 0.06\\
    \weighted & 27.80 & 8.16 \\
    \aware & 7.69 & 4.98 \\
    \hline
    \hline
        \textbf{Spanish}  & $J_{LTrain}$ & $J_{Test}$ \\
    \hline
    \uniform & 0.08 & 0.06\\
    \weighted & 27.81 & 8.07\\
    \aware & 6.89 & 4.65 \\
    \hline
    \hline
        \textbf{Swahili}  & $J_{LTrain}$ & $J_{Test}$ \\
    \hline
    \uniform & 3.06 & 3.74\\
    \weighted & 41.20 & 24.06\\
    \aware & 11.97 & 15.95\\
    \hline
    \hline
        \textbf{Turkish}  & $J_{LTrain}$ & $J_{Test}$ \\
    \hline
    \uniform & 0.10 & 0.11 \\
    \weighted & 27.91 & 7.66 \\
    \aware & 3.37 & 2.21 \\
    \end{tabular}
    \caption{Average Jaccard similarity    
    quantifying overlap between large training samples ($J_{LTrain}$) across random seeds and similarity between test samples ($J_{Test}$) across seeds. $J\in[0,100]$ where $100$ indicates that all UniMorph triples appear in all training sets}
    \label{tab:strategy-summaryjacc-bylang}
\end{table}

\begin{table}[h!]
    \centering
    \scriptsize
    \begin{tabular}{c|ccc|ccc}
    \multicolumn{1}{c}{} & \multicolumn{3}{c}{\textbf{Raw UniMorph}} & \multicolumn{3}{c}{\textbf{UniMorph$\times$Freq}}\\
    & \textbf{\#L} & \textbf{\#F} & \textbf{\#T} & \textbf{\#L} & \textbf{\#F} & \textbf{\#T} \\
    \hline
    \textbf{Arabic} & 12815 & 567 & 834113 & 11628 & 300 & 56035 \\
    \textbf{English} & 399758 & 11 & 716093 & 8370 & 6 & 16528 \\
    \textbf{German} & 39417 & 113 & 599141 & 4460 & 44 & 10501 \\
    \textbf{Spanish} & 65689 & 175 & 1286348 & 3592 & 117 & 11337 \\
    \textbf{Swahili} & 184 & 257 & 15149 & 180 & 225 & 3725 \\
    \textbf{Turkish} & 3579 & 883 & 570420 & 1649 & 242 & 24332 \\
    \end{tabular}
    \caption{Type frequencies for lemmas (\#L), feature sets (\#F), and triples (\#T) for each language data set. Raw UniMorph (3+)4 and intersected with frequency.}
    \label{tab:data-summary}
\end{table}

\begin{table*}[h!]
    \centering
    \scriptsize
    \begin{tabular}{l||cccc|cccc|cc|cc}
    \textbf{Overall Test vs S Train} & \both\% ($\sigma$) & \feats\ & \lemma\ & \neither & \featsAttested & \featsNovel\\
    \hline
    \uniform & 15.02 (\textit{25.29}) &  65.31  (\textit{33.2}) & 6.25 (\textit{8.32})  & 13.43 (\textit{14.57})  & 80.33 (\textit{19.50}) & 19.67 (\textit{19.50}) \\
    \weighted & 25.69  (\textit{15.61}) & 64.75  (\textit{25.01}) & 6.97  (\textit{10.67}) & 2.59  (\textit{2.42}) & 90.44  (\textit{11.13}) & 9.56  (\textit{11.13})\\
    \aware & 13.27  (\textit{13.43}) & 35.54  (\textit{13.96}) & 14.92  (\textit{15.20}) & 36.27  (\textit{14.71}) & 48.81  (\textit{0.98}) & 51.19  (\textit{0.98})\\
    \hline
    \hline
    \textbf{Overall Test vs L Train} & \both\% ($\sigma$) & \feats\ & \lemma\ & \neither & \featsAttested & \featsNovel\\
    \hline
    \uniform & 30.58 (\textit{32.47}) & 65.59 (\textit{35.62}) & 2.83 (\textit{4.56}) & 1.00 (\textit{}1.40) & 96.17 (\textit{5.55}) & 3.83 (\textit{5.55}) \\
    \weighted & 50.59 (\textit{16.38}) & 44.76 (\textit{21.74}) & 4.24 (\textit{7.22}) & 0.39 (\textit{0.58}) & 95.36 (\textit{7.28}) & 4.64 (\textit{7.28})\\
    \aware & 23.94 (\textit{14.76}) & 25.97 (\textit{14.84}) & 25.17 (\textit{14.14}) & 24.91 (\textit{14.05}) & 49.92 (\textit{0.17}) & 50.08 (\textit{0.17})\\
    \hline
    \hline
        \multicolumn{7}{c}{}\\
        \hline
    \hline
    \textbf{Ara Test vs STrain} & \both\% ($\sigma$) & \feats\ & \lemma\ & \neither & \featsAttested & \featsNovel\\
    \hline
\uniform & 
3.12 (\textit{0.26}) & 66.38 (\textit{4.22}) & 1.32 (\textit{0.35}) & 29.18 (\textit{4.02}) & 69.50 (\textit{4.14}) & 30.50 (\textit{4.14}) \\
\weighted & 
13.02 (\textit{1.18}) & 77.52 (\textit{1.33}) & 2.06 (\textit{0.40}) & 7.40 (\textit{1.14}) & 90.54 (\textit{1.53}) & 9.46 (\textit{1.53}) \\
\aware & 
3.06 (\textit{0.62}) & 44.62 (\textit{0.92}) & 3.30 (\textit{0.72}) & 49.02 (\textit{1.14}) & 47.68 (\textit{0.57}) & 52.32 (\textit{0.57}) \\
    \hline
    \hline
    \textbf{Ara Test vs LTrain} & \both\% ($\sigma$) & \feats\ & \lemma\ & \neither & \featsAttested & \featsNovel\\
    \hline
\uniform & 
15.82 (\textit{1.03}) & 80.82 (\textit{2.10}) & 0.78 (\textit{0.26}) & 2.58 (\textit{1.08}) & 96.64 (\textit{1.30}) & 3.36 (\textit{1.30}) \\
\weighted & 
39.38 (\textit{1.17}) & 57.42 (\textit{0.78}) & 1.66 (\textit{0.62}) & 1.54 (\textit{0.46}) & 96.80 (\textit{0.77}) & 3.20 (\textit{0.77}) \\
\aware & 
10.40 (\textit{1.31}) & 39.50 (\textit{1.24}) & 10.82 (\textit{0.84}) & 39.28 (\textit{0.86}) & 49.90 (\textit{0.11}) & 50.10 (\textit{0.11}) \\

    \hline
    \multicolumn{7}{c}{}\\
    \hline
    \textbf{Deu Test vs STrain} & \both\% ($\sigma$) & \feats\ & \lemma\ & \neither & \featsAttested & \featsNovel\\
    \hline
\uniform & 
1.16 (\textit{0.52}) & 97.42 (\textit{1.09}) & 0.00 (\textit{0.00}) & 1.42 (\textit{0.84}) & 98.58 (\textit{0.84}) & 1.42 (\textit{0.84}) \\
\weighted & 
12.08 (\textit{0.50}) & 85.90 (\textit{1.34}) & 0.74 (\textit{0.43}) & 1.28 (\textit{0.70}) & 97.98 (\textit{1.11}) & 2.02 (\textit{1.11}) \\
\aware & 
4.70 (\textit{1.40}) & 45.20 (\textit{1.50}) & 4.90 (\textit{1.13}) & 45.20 (\textit{1.19}) & 49.90 (\textit{0.15}) & 50.10 (\textit{0.15}) \\
    \hline
    \hline
    \textbf{Deu Test vs LTrain} & \both\% ($\sigma$) & \feats\ & \lemma\ & \neither & \featsAttested & \featsNovel\\
    \hline
\uniform & 
4.38 (\textit{0.34}) & 95.42 (\textit{0.43}) & 0.00 (\textit{0.00}) & 0.20 (\textit{0.13}) & 99.80 (\textit{0.13}) & 0.20 (\textit{0.13}) \\
\weighted & 
36.38 (\textit{1.24}) & 63.50 (\textit{1.24}) & 0.08 (\textit{0.07}) & 0.04 (\textit{0.05}) & 99.88 (\textit{0.10}) & 0.12 (\textit{0.10}) \\
\aware & 
14.74 (\textit{3.32}) & 35.26 (\textit{3.32}) & 14.96 (\textit{2.28}) & 35.04 (\textit{2.28}) & 50.00 (\textit{0.00}) & 50.00 (\textit{0.00}) \\
        \hline
    \multicolumn{7}{c}{}\\
    \hline
    \textbf{Eng Test vs STrain} & \both\% ($\sigma$) & \feats\ & \lemma\ & \neither & \featsAttested & \featsNovel\\
    \hline
\uniform & 
0.10 (\textit{0.11}) & 99.68 (\textit{0.26}) & 0.00 (\textit{0.00}) & 0.22 (\textit{0.29}) & 99.78 (\textit{0.29}) & 0.22 (\textit{0.29}) \\
\weighted & 
10.62 (\textit{0.82}) & 89.38 (\textit{0.82}) & 0.00 (\textit{0.00}) & 0.00 (\textit{0.00}) & 100.00 (\textit{0.00}) & 0.00 (\textit{0.00}) \\
\aware & 
1.94 (\textit{0.61}) & 48.06 (\textit{0.61}) & 3.02 (\textit{0.39}) & 46.98 (\textit{0.39}) & 50.00 (\textit{0.00}) & 50.00 (\textit{0.00}) \\
    \hline
    \hline
    \textbf{Eng Test vs LTrain} & \both\% ($\sigma$) & \feats\ & \lemma\ & \neither & \featsAttested & \featsNovel\\
    \hline
\uniform & 
0.38 (\textit{0.07}) & 99.62 (\textit{0.07}) & 0.00 (\textit{0.00}) & 0.00 (\textit{0.00}) & 100.00 (\textit{0.00}) & 0.00 (\textit{0.00}) \\
\weighted & 
31.26 (\textit{0.91}) & 68.74 (\textit{0.91}) & 0.00 (\textit{0.00}) & 0.00 (\textit{0.00}) & 100.00 (\textit{0.00}) & 0.00 (\textit{0.00}) \\
\aware & 
7.16 (\textit{2.48}) & 42.84 (\textit{2.48}) & 12.04 (\textit{0.63}) & 37.96 (\textit{0.63}) & 50.00 (\textit{0.00}) & 50.00 (\textit{0.00}) \\
        \hline
    \multicolumn{7}{c}{}\\
    \hline
    \textbf{Spa Test vs STrain} & \both\% ($\sigma$) & \feats\ & \lemma\ & \neither & \featsAttested & \featsNovel\\
    \hline
\uniform & 
3.88 (\textit{0.56}) & 84.60 (\textit{2.48}) & 0.46 (\textit{0.22}) & 11.06 (\textit{1.99}) & 88.48 (\textit{2.14}) & 11.52 (\textit{2.14}) \\
\weighted & 
28.40 (\textit{1.40}) & 63.54 (\textit{1.78}) & 5.94 (\textit{0.71}) & 2.12 (\textit{0.67}) & 91.94 (\textit{1.12}) & 8.06 (\textit{1.12}) \\
\aware & 
15.02 (\textit{3.78}) & 34.00 (\textit{3.71}) & 15.54 (\textit{2.03}) & 35.44 (\textit{1.94}) & 49.02 (\textit{0.17}) & 50.98 (\textit{0.17}) \\
    \hline
    \hline
    \textbf{Spa Test vs LTrain} & \both\% ($\sigma$) & \feats\ & \lemma\ & \neither & \featsAttested & \featsNovel\\
    \hline
\uniform & 
16.72 (\textit{0.61}) & 83.28 (\textit{0.61}) & 0.00 (\textit{0.00}) & 0.00 (\textit{0.00}) & 100.00 (\textit{0.00}) & 0.00 (\textit{0.00}) \\
\weighted & 
53.30 (\textit{1.58}) & 44.76 (\textit{1.64}) & 1.74 (\textit{0.49}) & 0.20 (\textit{0.23}) & 98.06 (\textit{0.69}) & 1.94 (\textit{0.69}) \\
\aware & 
28.08 (\textit{4.52}) & 21.90 (\textit{4.53}) & 28.02 (\textit{4.10}) & 22.00 (\textit{4.10}) & 49.98 (\textit{0.04}) & 50.02 (\textit{0.04}) \\
        \hline
    \multicolumn{7}{c}{}\\
    \hline
    \textbf{Swc Test vs STrain} & \both\% ($\sigma$) & \feats\ & \lemma\ & \neither & \featsAttested & \featsNovel\\
    \hline
\uniform & 
70.98 (\textit{2.51}) & 11.12 (\textit{2.25}) & 16.02 (\textit{1.50}) & 1.88 (\textit{0.50}) & 82.10 (\textit{1.76}) & 17.90 (\textit{1.76}) \\
\weighted & 
52.24 (\textit{5.04}) & 15.10 (\textit{1.39}) & 30.00 (\textit{4.90}) & 2.66 (\textit{0.94}) & 67.34 (\textit{5.76}) & 32.66 (\textit{5.76}) \\
\aware & 
40.68 (\textit{1.10}) & 7.04 (\textit{1.07}) & 46.52 (\textit{1.33}) & 5.76 (\textit{1.41}) & 47.72 (\textit{0.30}) & 52.28 (\textit{0.30}) \\
    \hline
    \hline
    \textbf{Swc Test vs LTrain} & \both\% ($\sigma$) & \feats\ & \lemma\ & \neither & \featsAttested & \featsNovel\\
    \hline
\uniform & 
91.82 (\textit{0.65}) & 4.34 (\textit{0.77}) & 3.66 (\textit{0.53}) & 0.18 (\textit{0.16}) & 96.16 (\textit{0.63}) & 3.84 (\textit{0.63}) \\
\weighted & 
72.62 (\textit{2.51}) & 6.86 (\textit{1.18}) & 20.12 (\textit{2.43}) & 0.40 (\textit{0.22}) & 79.48 (\textit{2.63}) & 20.52 (\textit{2.63}) \\
\aware & 
47.64 (\textit{1.07}) & 2.04 (\textit{1.08}) & 48.70 (\textit{0.98}) & 1.62 (\textit{0.89}) & 49.68 (\textit{0.29}) & 50.32 (\textit{0.29}) \\
        \hline
    \multicolumn{7}{c}{}\\
    \hline
    \textbf{Tur Test vs STrain} & \both\% ($\sigma$) & \feats\ & \lemma\ & \neither & \featsAttested & \featsNovel\\
    \hline
\uniform & 
10.88 (\textit{0.63}) & 32.64 (\textit{2.15}) & 19.68 (\textit{0.90}) & 36.80 (\textit{1.16}) & 43.52 (\textit{1.88}) & 56.48 (\textit{1.88}) \\
\weighted & 
37.80 (\textit{1.51}) & 57.06 (\textit{1.13}) & 3.06 (\textit{0.78}) & 2.08 (\textit{0.41}) & 94.86 (\textit{1.03}) & 5.14 (\textit{1.03}) \\
\aware & 
14.24 (\textit{1.67}) & 34.30 (\textit{1.45}) & 16.22 (\textit{0.72}) & 35.24 (\textit{0.66}) & 48.54 (\textit{0.28}) & 51.46 (\textit{0.28}) \\
    \hline
    \hline
    \textbf{Tur Test vs LTrain} & \both\% ($\sigma$) & \feats\ & \lemma\ & \neither & \featsAttested & \featsNovel\\
    \hline
\uniform & 
54.36 (\textit{0.81}) & 30.06 (\textit{0.75}) & 12.52 (\textit{1.21}) & 3.06 (\textit{0.72}) & 84.42 (\textit{1.35}) & 15.58 (\textit{1.35}) \\
\weighted & 
70.62 (\textit{1.33}) & 27.30 (\textit{1.26}) & 1.88 (\textit{0.61}) & 0.20 (\textit{0.11}) & 97.92 (\textit{0.53}) & 2.08 (\textit{0.53}) \\
\aware & 
35.64 (\textit{1.06}) & 14.30 (\textit{1.04}) & 36.50 (\textit{1.52}) & 13.56 (\textit{1.47}) & 49.94 (\textit{0.08}) & 50.06 (\textit{0.08}) \\
    \end{tabular}
    
    \caption{Language-by-language average mean percentage of each overlap type in test sets relative to small and large training. Standard deviations are (\textit{italicized}). \aware\ targets a \featsAttested\ relative to large train as close to 50\% as possible without exceeding it. \%\featsAttested\ = \%\both\ + \%\feats\ and \%\featsNovel\ = \%\lemma\ + \%\neither.}
    \label{tab:strategy-summary2-bylang}
\end{table*}

\newpage

\pagebreak

\section{Detailed Results}
\label{sec:appendix-results}

This appendix contains Tables \ref{tab:strategy-summaryres-bymodel}-\ref{tab:strategy-summaryres-bylang}.

\begin{table*}[h!]
    \centering
    \scriptsize
    \begin{tabular}{l||cccc|cc|cc|c}
\textbf{\nonneural\ Test vs S Train} & \both\% & \feats\ & \lemma\ & \neither & \featsAttested & \featsNovel & overall\\
\hline
\uniform &
70.92 & 66.75 & 17.16 & 19.10 & 67.50 & 16.94 & 59.83 \\
\weighted &
67.86 & 77.93 & 8.15 & 13.07 & 74.98 & 9.91 & 68.79 \\
\aware &
66.47 & 75.43 & 17.79 & 26.55 & 73.39 & 24.63 & 48.30 \\
\hline
\textbf{\nonneural\ Test vs L Train} & \both\% & \feats\ & \lemma\ & \neither & \featsAttested & \featsNovel & overall\\
\hline
\uniform &
73.59 & 66.00 & 21.85 & 25.75 & 71.66 & 31.72 & 70.33 \\
\weighted &
75.35 & 83.62 & 8.06 & 9.17 & 79.15 & 7.61 & 76.1o \\
\aware &
74.52 & 82.49 & 18.57 & 29.31 & 77.84 & 24.33 & 51.03 \\
\hline
\multicolumn{8}{c}{}\\
\hline

\textbf{\wuetal\ Test vs S Train} & \both\% & \feats\ & \lemma\ & \neither & \featsAttested & \featsNovel & overall\\
\hline
\uniform &
70.02 & 61.05 & 58.61 & 30.48 & 67.70 & 39.36 & 65.33 \\
\weighted &
79.18 & 69.36 & 43.60 & 26.20 & 75.08 & 36.15 & 72.27 \\
\aware &
80.28 & 72.46 & 38.15 & 30.86 & 78.06 & 35.97 & 56.67 \\
\hline
\textbf{\wuetal\ Test vs L Train} & \both\% & \feats\ & \lemma\ & \neither & \featsAttested & \featsNovel & overall\\
\hline
\uniform &
79.60 & 76.61 & 63.85 & 39.92 & 79.51 & 55.72 & 78.82 \\
\weighted &
89.42 & 85.42 & 59.62 & 37.81 & 89.48 & 52.64 & 88.56 \\
\aware &
89.78 & 86.56 & 45.65 & 38.87 & 89.83 & 43.92 & 66.85 \\
\hline
\multicolumn{8}{c}{}\\
\hline

\textbf{\cluzhbfour\ Test vs S Train} & \both\% & \feats\ & \lemma\ & \neither & \featsAttested & \featsNovel & overall\\
\hline
\uniform &
77.09 & 71.75 & 57.13 & 33.22 & 73.87 & 39.72 & 70.29 \\
\weighted &
78.35 & 86.22 & 26.18 & 21.40 & 83.67 & 22.63 & 78.09 \\
\aware &
79.97 & 84.86 & 30.43 & 32.00 & 83.66 & 32.16 & 57.38 \\
\hline
\textbf{\cluzhbfour\ Test vs L Train} & \both\% & \feats\ & \lemma\ & \neither & \featsAttested & \featsNovel & overall\\
\hline
\uniform &
88.14 & 79.80 & 72.66 & 47.34 & 86.02 & 69.86 & 85.42 \\
\weighted &
86.14 & 90.39 & 20.63 & 20.93 & 88.22 & 17.71 & 85.83 \\
\aware &
88.31 & 91.81 & 35.35 & 41.20 & 89.78 & 37.68 & 63.70 \\
\hline
\multicolumn{8}{c}{}\\
\hline

\textbf{\cluzhgr\ Test vs S Train} & \both\% & \feats\ & \lemma\ & \neither & \featsAttested & \featsNovel & overall\\
\hline
\uniform &
75.72 & 70.77 & 55.27 & 31.89 & 72.83 & 38.27 & 69.21 \\
\weighted &
77.79 & 85.91 & 25.75 & 21.22 & 83.28 & 22.38 & 77.72 \\
\aware &
79.78 & 84.50 & 29.98 & 31.49 & 83.28 & 31.78 & 57.00 \\
\hline
\textbf{\cluzhgr\ Test vs L Train} & \both\% & \feats\ & \lemma\ & \neither & \featsAttested & \featsNovel & overall\\
\hline
\uniform &
85.15 & 75.83 & 65.54 & 43.43 & 82.83 & 65.00 & 82.24 \\
\weighted &
84.65 & 89.17 & 20.17 & 17.13 & 86.89 & 17.01 & 84.52 \\
\aware &
85.76 & 89.64 & 33.91 & 40.04 & 87.42 & 36.12 & 61.74 \\
       \end{tabular}
    
    \caption{Average percent accuracy across seeds and models on the test set by architecture.}
    \label{tab:strategy-summaryres-bymodel}
\end{table*}

\begin{table*}[h!]
    \centering
    \scriptsize
    \begin{tabular}{l||cccc|cc|cc|c}
\textbf{Overall Test vs S Train} & \both\% & \feats\ & \lemma\ & \neither & \featsAttested & \featsNovel & overall\\
\hline
\uniform &
73.44 & 67.58 & 47.05 & 28.67 & 70.47 & 33.57 & 66.16 \\
\weighted &
75.79 & 79.86 & 25.92 & 20.47 & 79.25 & 22.77 & 74.22 \\
\aware &
76.62 & 79.31 & 29.09 & 30.22 & 79.60 & 31.13 & 54.84 \\
\hline
\textbf{Overall Test vs L Train} & \both\% & \feats\ & \lemma\ & \neither & \featsAttested & \featsNovel & overall\\
\hline
\uniform &
81.62 & 74.56 & 55.97 & 39.11 & 80.00 & 55.57 & 79.20 \\
\weighted &
83.89 & 87.15 & 27.12 & 21.26 & 85.94 & 23.74 & 83.75 \\
\aware &
84.59 & 87.63 & 33.37 & 37.36 & 86.22 & 35.51 & 60.83 \\
\hline
\hline
\multicolumn{8}{c}{}\\
\hline
\hline
\textbf{Ara Test vs S Train} & \both\% & \feats\ & \lemma\ & \neither & \featsAttested & \featsNovel & overall\\
\hline
\uniform &
72.52 & 67.86 & 54.84 & 50.58 & 68.06 & 50.80 & 62.80 \\
\weighted &
73.82 & 73.15 & 35.79 & 23.98 & 73.24 & 26.54 & 68.82 \\
\aware &
63.77 & 66.33 & 33.42 & 30.97 & 66.14 & 31.11 & 47.81 \\
\hline
\textbf{Ara Test vs L Train} & \both\% & \feats\ & \lemma\ & \neither & \featsAttested & \featsNovel & overall\\
\hline
\uniform &
83.60 & 76.52 & 62.57 & 44.31 & 77.67 & 48.62 & 76.76 \\
\weighted &
79.92 & 78.95 & 38.29 & 23.67 & 79.34 & 31.04 & 77.76 \\
\aware &
75.07 & 76.36 & 46.49 & 45.99 & 76.09 & 46.09 & 61.06 \\
\hline
\multicolumn{8}{c}{}\\
\hline

\textbf{Deu Test vs S Train} & \both\% & \feats\ & \lemma\ & \neither & \featsAttested & \featsNovel & overall\\
\hline
\uniform &
63.61 & 60.00 & \--- & 28.27 & 60.06 & 28.27 & 59.65 \\
\weighted &
78.22 & 76.73 & 26.06 & 16.48 & 76.91 & 20.18 & 75.81 \\
\aware &
73.90 & 73.88 & 38.98 & 41.80 & 74.12 & 41.60 & 57.84 \\
\hline
\textbf{Deu Test vs L Train} & \both\% & \feats\ & \lemma\ & \neither & \featsAttested & \featsNovel & overall\\
\hline
\uniform &
75.37 & 73.07 & \--- & 73.33 & 73.16 & 73.33 & 73.14 \\
\weighted &
85.35 & 84.37 & 25.00 & 0.00 & 84.72 & 14.58 & 84.64 \\
\aware &
81.22 & 82.00 & 40.02 & 44.25 & 81.84 & 43.24 & 62.54 \\
\hline
\multicolumn{8}{c}{}\\
\hline

\textbf{Eng Test vs S Train} & \both\% & \feats\ & \lemma\ & \neither & \featsAttested & \featsNovel & overall\\
\hline
\uniform &
97.22 & 93.34 & \--- & 0.00 & 93.35 & 0.00 & 93.14 \\
\weighted &
76.90 & 88.43 & \--- & \--- & 87.20 & \--- & 87.20 \\
\aware &
84.30 & 88.53 & 17.10 & 19.14 & 88.45 & 18.99 & 53.72 \\
\hline
\textbf{Eng Test vs L Train} & \both\% & \feats\ & \lemma\ & \neither & \featsAttested & \featsNovel & overall\\
\hline
\uniform &
95.66 & 96.49 & \--- & \--- & 96.48 & \--- & 96.48 \\
\weighted &
84.25 & 95.26 & \--- & \--- & 91.83 & \--- & 91.83 \\
\aware &
89.96 & 92.11 & 17.81 & 19.80 & 91.95 & 19.32 & 55.63 \\
\hline
\multicolumn{8}{c}{}\\
\hline

\textbf{Spa Test vs S Train} & \both\% & \feats\ & \lemma\ & \neither & \featsAttested & \featsNovel & overall\\
\hline
\uniform &
75.09 & 71.24 & 46.87 & 39.58 & 71.35 & 39.67 & 67.67 \\
\weighted &
65.97 & 83.03 & 10.02 & 8.36 & 77.74 & 9.59 & 72.22 \\
\aware &
68.60 & 84.40 & 9.94 & 27.14 & 79.90 & 21.92 & 50.35 \\
\hline
\textbf{Spa Test vs L Train} & \both\% & \feats\ & \lemma\ & \neither & \featsAttested & \featsNovel & overall\\
\hline
\uniform &
84.09 & 83.39 & \--- & \--- & 83.50 & \--- & 83.50 \\
\weighted &
80.73 & 92.16 & 24.60 & 38.89 & 85.94 & 24.74 & 84.77 \\
\aware &
82.57 & 94.20 & 16.06 & 35.42 & 87.92 & 24.83 & 56.37 \\
\hline
\multicolumn{8}{c}{}\\
\hline

\textbf{Swc Test vs S Train} & \both\% & \feats\ & \lemma\ & \neither & \featsAttested & \featsNovel & overall\\
\hline
\uniform &
89.68 & 69.89 & 63.61 & 31.14 & 87.02 & 60.08 & 82.22 \\
\weighted &
80.41 & 75.56 & 29.41 & 26.04 & 79.27 & 29.12 & 62.79 \\
\aware &
85.83 & 78.31 & 43.16 & 31.05 & 84.79 & 41.75 & 62.28 \\
\hline
\textbf{Swc Test vs L Train} & \both\% & \feats\ & \lemma\ & \neither & \featsAttested & \featsNovel & overall\\
\hline
\uniform &
90.74 & 58.56 & 59.70 & 6.25 & 89.26 & 57.27 & 88.01 \\
\weighted &
82.30 & 77.40 & 40.77 & 33.75 & 81.88 & 40.66 & 73.36 \\
\aware &
88.53 & 88.42 & 44.11 & 43.24 & 88.56 & 44.01 & 66.14 \\
\hline
\multicolumn{8}{c}{}\\
\hline

\textbf{Tur Test vs S Train} & \both\% & \feats\ & \lemma\ & \neither & \featsAttested & \featsNovel & overall\\
\hline
\uniform &
42.51 & 43.14 & 22.85 & 22.46 & 42.99 & 22.61 & 31.51 \\
\weighted &
79.46 & 82.24 & 28.32 & 27.51 & 81.15 & 28.41 & 78.46 \\
\aware &
83.33 & 84.42 & 31.93 & 31.23 & 84.18 & 31.43 & 57.03 \\
\hline
\textbf{Tur Test vs L Train} & \both\% & \feats\ & \lemma\ & \neither & \featsAttested & \featsNovel & overall\\
\hline
\uniform &
60.24 & 59.34 & 45.65 & 32.55 & 59.94 & 43.08 & 57.33 \\
\weighted &
90.80 & 94.75 & 6.93 & 10.00 & 91.91 & 7.70 & 90.16 \\
\aware &
90.21 & 92.67 & 35.72 & 35.44 & 90.94 & 35.59 & 63.23 \\
    \end{tabular}
    
    \caption{Language-by-language average percent accuracy across seeds and models on the test set. Dashes indicate overlap partitions with size zero.}
    \label{tab:strategy-summaryres-bylang}
\end{table*}

\end{document}